\documentclass[runningheads]{llncs}
\usepackage{graphicx}
\usepackage[section]{placeins}

\usepackage{multirow} 
\usepackage{rotating} 
\usepackage{bigstrut} 
\usepackage{booktabs}
\usepackage{tabularx}
\usepackage{array}
\newcolumntype{P}[1]{>{\centering\arraybackslash}p{#1}}
\usepackage[english]{babel}

\usepackage[dvipsnames]{xcolor}

\usepackage[inkscapeformat=png]{svg}
\newcommand{\ourdataset}{\textsc{WD-Known}}

\begin{document}
\title{Evaluating Language Models\\ for  Knowledge Base Completion}
\vspace{-3mm}


\author{Blerta Veseli\inst{1} \and
Sneha Singhania\inst{1}
\and
Simon Razniewski\inst{2}
\and
Gerhard Weikum\inst{1}}

\authorrunning{Veseli, Singhania, Razniewski and Weikum}
\institute{Max Planck Institute for Informatics \and
Bosch Center for AI}
%
\maketitle  


\begin{abstract}

Structured knowledge bases (KBs) are a foundation of many intelligent applications, yet are notoriously incomplete.
Language models (LMs) have recently been proposed for unsupervised knowledge base completion (KBC), yet, despite encouraging initial results, questions regarding their suitability remain open.
Existing evaluations often fall short because they only evaluate on popular subjects, or sample already existing facts from KBs.
In this work, we introduce a novel, more challenging benchmark dataset, and a methodology tailored for a realistic assessment of the KBC potential of LMs. 
For automated assessment, we curate a dataset called \ourdataset{}, which provides an unbiased random sample of Wikidata, containing over 3.9 million facts.
In a second step, we perform a human evaluation on predictions that are not yet in the KB, as only this provides real insights into the added value over existing KBs.
Our key finding is that biases in dataset conception of previous benchmarks lead to a systematic overestimate of LM performance for KBC. %
However, our results also reveal strong areas of LMs.
We could, for example, perform a significant completion of Wikidata on the relations \texttt{nativeLanguage}, by a factor of $\sim21$  (from 260k to 5.8M) at 82\% precision, \texttt{usedLanguage}, by a factor of $\sim 2.1$ (from 2.1M to 6.6M) at 82\% precision,
and \texttt{citizenOf} by a factor of $\sim0.3$ (from 4.2M to 5.3M) at 90\% precision.
Moreover, we find that LMs possess surprisingly strong generalization capabilities: even on relations where most facts were not directly observed in LM training, prediction quality can be high. We open-source the benchmark dataset and code.\footnote{https://github.com/bveseli/LMsForKBC}

\end{abstract}
\section{Introduction}
\label{intro}

Structured knowledge bases (KBs) like Wikidata~\cite{wikidata}, DBpedia~\cite{dbpedia}, and Yago~\cite{yago} are backbones of the semantic web and are employed in many knowledge-centric applications like search, question answering and dialogue. Constructing these KBs at high quality and scale is a long-standing research challenge and multiple knowledge base construction benchmarks exist, e.g., FB15k~\cite{bordes2013translating}, CoDEx~\cite{safavi2020codex}, and LM-KBC22~\cite{singhania2022lm}. Text-extraction, knowledge graph embeddings, and LM-based knowledge extraction have continuously moved scores upwards on these tasks, and leaderboard portals like Paperswithcode\footnote{\url{https://paperswithcode.com/task/knowledge-graph-completion}} provide evidence to that.

Recently, pre-trained language models have been purported as a promising source for structured knowledge. Starting from the seminal LAMA paper \cite{petroni}, a throve of works have explored how to better probe, train, or fine-tune these LMs \cite{liu2021pre}. Nonetheless, we observe a certain divide between these late-breaking investigations, and practical knowledge base completion (KBC). While the recent LM-based approaches often focus on attractive and clean methodologies that produce fast results, practical KBC is a highly precision-oriented, extremely laborious process, involving a very high degree of manual labor, either for manually creating statements \cite{wikidata}, or for building comprehensive scraping, cleaning, validation, and normalization pipelines \cite{dbpedia,yago}. We believe previous works fall short in three aspects:

\begin{enumerate}
    \item \textit{Focus on high precision:} On the KB side, part of Yago's success stems from its validated $>$95\% accuracy, and the Google Knowledge Vault was not deployed into production, partly because it did not achieve 99\% accuracy \cite{weikum-machine-knowledge}. Yet, previous LM analyses balance precision and recall or report precision/hits@k values, implicitly tuning systems towards balanced recall scores resulting in impractical precision.
    \item \textit{Evaluation of completion potential:} Existing benchmarks often sample from popular subjects. This is useful for system comparison, but not for KBC. E.g., predicting capitals of countries with 99\% accuracy does not imply practical value: they are already captured in established KBs like Wikidata.
    \item \textit{Prediction of missing facts:} Existing works test LMs on facts already stored in KBs, which does not provide us with a realistic assessment for completion. For KBC we need to predict objects for subject-relation pairs, previously not known to the KB. 
\end{enumerate}
It is also important to keep in mind the scale of KBs: Wikidata, for instance, currently contains around 100 Million entities, and 1.2B statements. The cost of producing such KBs is massive, one estimate from 2018 sets the cost per triple at 2 USD for manually curated triples\cite{paulheim}, and 1 ct for automatically extracted ones.\footnote{Wikidata might broadly fall in between, as its aim is human-curated quality, but major portions are imported semi-automatically from other sources.} Thus, even small additions in relative terms, might correspond to massive gains in absolute numbers. For example, even by the lower estimate of 1 ct/triple, adding one triple to just 1\% of Wikidata humans, would come at a cost of 100k USD.

In this paper, \textit{we conduct a systematic analysis of LMs for KBC}, where we focus on \textit{high precision ranges} (90\%). We evaluate by first using a new benchmark dataset \ourdataset{}, where we randomly sample facts from Wikidata (including many without any structured information, Wikipedia information, or English labels) and second by a manual evaluation on subject-relation pairs without object values, yet.

Technically, we focus on the BERT language model \cite{bert}, and the Wikidata knowledge base. Although BERT has been superseded by newer LMs, its popularity is still matched only by the closed source GPT-3 and chatGPT models. Wikidata is by far the most prominent and comprehensive public KB, so evaluations against it provide the strongest yardstick.

Our main results are as follows:
\begin{enumerate}
    \item In actual KBC, LMs perform considerably worse than benchmarks like LAMA indicated, but still achieve strong results for language-related, socio-demographic relations (e.g., \textit{nativeLanguage}).
    \item Simple changes on out-of-the-box LMs, in particular, vocabulary expansion and prompt formulation, can significantly improve their ability to output high-precision knowledge.
    \item Using LMs, Wikidata could be significantly expanded for three relations, \texttt{native\-Language} by a factor of $\sim21$ (from 260k to 5.8M) at precision $82\%$, \texttt{citizenOf} by a factor of $\sim0.3$ (from 4.2M to 5.3M) at 90\% precision and \texttt{usedLanguage} by a factor of $\sim2.08$ (from 2.1M to 6.6M) at 82\% precision.
\end{enumerate}

\section{Background and Related Work}

\paragraph{KB construction and completion}

Knowledge base construction is a field with considerable history. One prominent approach is by human curation, as done e.g., in the seminal CYC project \cite{lenat1995cyc}, and this is also the backbone of today's most prominent public KB, Wikidata \cite{wikidata}.
Another popular paradigm is the extraction from semi-structured resources, as pursued in Yago and DBpedia \cite{yago,dbpedia}. Extraction from free text has also been explored (e.g., NELL \cite{nell}). 
A popular paradigm has been embedding-based link prediction, e.g., via tensor factorization like Rescal \cite{nickel2011three}, and KG embeddings like TransE \cite{bordes2013translating}. 

An inherent design decision in KBC is the P/R trade-off -- academic projects are often open to trade these freely (e.g., via F-1 scores), while production environments are often very critically concerned with precision, e.g., Wikidata generally discouraging statistical inferences\footnote{There is often a terminological confusion here: Automated editing is omnipresent on Wikidata, but the bots performing them typically execute meticulously pre-defined edit and insertion tasks (e.g., based on other structured sources), not based on statistical inference.}, and industrial players likely using to a considerable degree human editing and verification \cite{weikum-machine-knowledge}.
Although the P/R trade-off is in principle tunable via thresholds, the high-precision range is hardly investigated. For example in all of Rescal, TransE, and LAMA, the main results focus on metrics like hits@k, MRR, or AUC, which provide no bounds on precision.

\paragraph{LMs for KB construction}

Knowledge extraction from language models provides fresh hope for the synergy of automated approaches and high-precision curated KBs. Knowledge-extraction from LMs provides remarkably straightforward access to very large text corpora: The basic idea by \cite{petroni} is to just define one template per relation, then query the LM with subject-instantiated versions, and retain its top prediction(s). A range of follow-up works appeared, focusing, e.g., on investigating entities, improving updates, exploring storage limits, incorporating unique entity identifiers, and others \cite{autoprompt,poerner2020ebert,decao2021editing,roberts2020knowledge,heinzerling-inui-2021-language,petroni2020context,elazar2021measuring,razniewski2021language,cohen2023crawling}. Nonetheless, we observe the same gaps as above: The high-precision area, and real KBC, are underexplored.

\paragraph{Benchmarking KB completion}

KB completion (sometimes also referred to as link prediction) has a set of quasi-standard benchmarks. Here we review the important ones and outline why they do not help for the focus of our investigation.


Two of the most popular benchmarks, both introduced in \cite{bordes2013translating}, are \textbf{FB15k} and \textbf{WN18}. The former contains statements for 15k extremely popular entities from Freebase, entities that already in 2013, when KBs were small, had at least 100 statements. The latter contains 41k entities from WordNet, yet these are linguistic concepts where WordNet is already the authoritative reference, and the potential for completion is small.
\textbf{DBP-5L} is a popular multilingual dataset of 14k entities, yet it is collected by iterative graph neighbourhood expansion, so by design is biased towards popular (well-connected) entities. \textbf{YAGO3-10} is another benchmark, that contains YAGO3 entities with at least 10 statements \cite{dettmers2018convolutional}.
The recent \textbf{CoDEx} benchmark provides a much larger subset of Wikidata triples but again focuses on the more popular subjects, as even its hardest variant considers only entities with at least 5 statements \cite{safavi2020codex}. The
\textbf{LAMA} benchmark \cite{petroni2020context} is based on the T-REx dataset, which in turn restricts the scope of subjects to those having a Wikipedia page. \textbf{LAMA-UHN} \cite{poerner2020ebert} removes some surface correlations but does not remove the restriction to Wikipedia-known subjects. \textbf{LM-KBC22} \cite{singhania2022lm} provides a curated mix of popular and long-tail subjects, but not a random sample, and only a small set of 12 relations. In summary, all these benchmarks provide non-random subject sets, and by taking ground truth from existing KBs, necessarily can not evaluate a method's real KB completion potential. The \textbf{PKGC} work \cite{pkgc} uses human evaluation to account for KB incompleteness, but also uses subjects from previous benchmarks focused on popular entities.

\section{Method}

\subsection{Knowledge Base Completion Tasks}

Established KBs like Wikidata store a large number of facts of the form \textit{(subject, relation, object)}. However, such KBs still suffer from incompleteness, which limits their applicability. KBC tries to counteract this incompleteness and describes the task of predicting missing facts for a KB. KBC is often split into subtasks, such as predicting the relation, subject, or object slots in triples. In the following, we focus on the most prominent object slot filling task, i.e. predicting an object for a subject-relation pair without an object so far. Identifying plausible subject-relation pairs is another important task, as not every combination qualifies, e.g., (Albert Einstein, hasCapital,$\cdot$ ) has no object.




We will refer to facts that are present in a KB as \textit{existing facts} and \textit{existing-fact prediction} describes predicting the object for a subject-relation pair for which the object value already exists in the KB. Similarly, we refer to facts that are missing in a KB as \textit{missing facts} and \textit{missing-fact prediction} describes predicting the object for a subject-relation pair with no object value yet.

\subsection{Fact Prediction using Pre-trained Language Models}
\label{model}

    \begin{figure}[!ht]
      \centering
      \includegraphics[width=115mm]{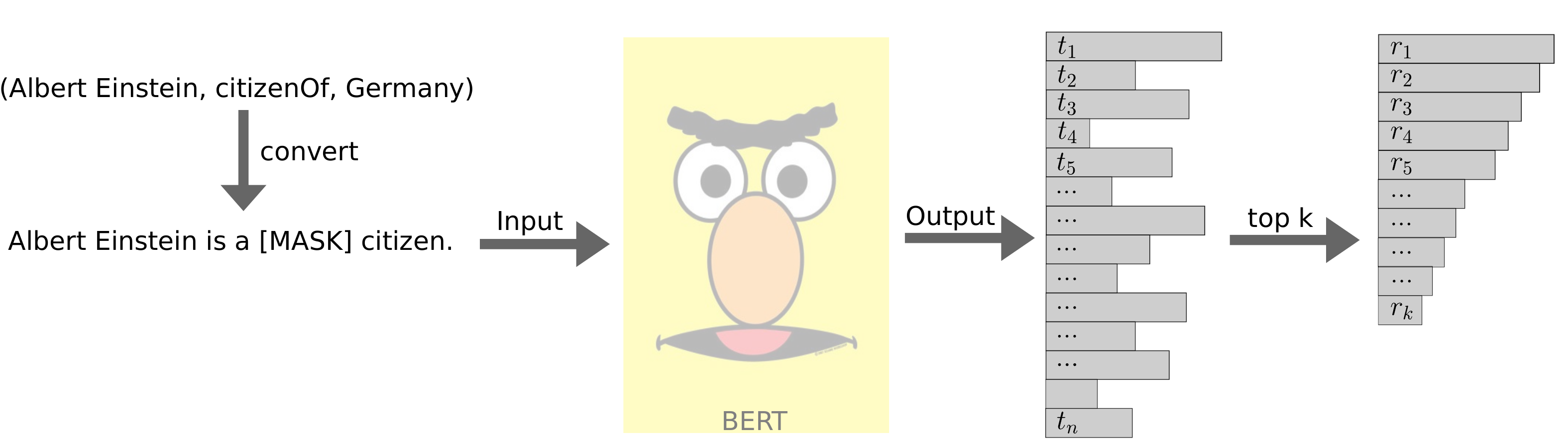}
      \caption{To query the LM for an object slot, we convert triples into relation-specific prompts by masking the object, following \cite{petroni}. The output is a probability vector over the model's vocabulary. We retain the top $k=10$ predictions.
      }
      \label{fig:model}
    \end{figure}

\noindent
The slot filling ability of an LM, i.e. predicting ``Paris`` for a pair (France, hasCapital), is essential for KBC. This is done by querying an LM using cloze-style statements, a.k.a. prompts, like ``The capital of France is [MASK].''\cite{petroni} . The LM's goal is to predict the token at the masked position, i.e. the object. 

We probe the masked language model BERT \cite{bert} and query for facts of different relations by using relation-specific predefined prompts. The prompts contain placeholders for subject and object, so that the input for the LM can be automatically created by replacing the subject placeholder with the actual subject and the object placeholder with [MASK]. For each cloze-style query like ``The capital of France is [MASK].'', BERT returns a probability vector over its vocabulary $[t_1,..,t_n]$ ($\sim$29K tokens). From this vector, we select \textit{top k} predictions $[r_1,..,r_k]$ with the highest probability as predictions for the object. We set $k=10$.

\begin{figure}
  \centering
  \includegraphics[width=120mm]{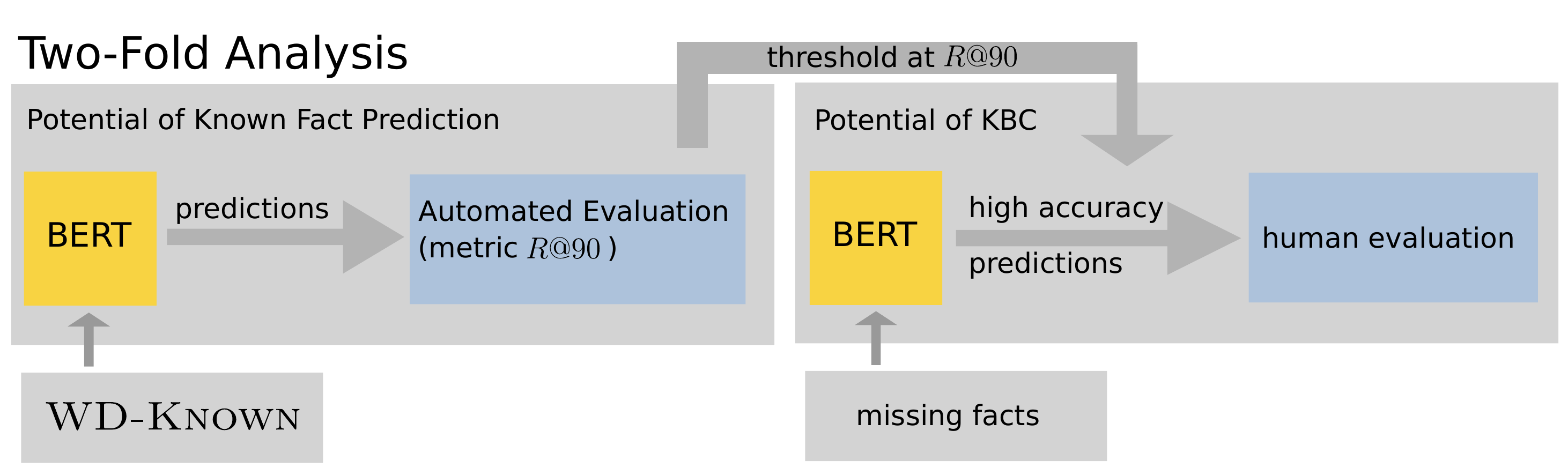}
  \caption{Our systematic analysis is divided into two parts. First, we analyse existing-fact prediction by an automated evaluation computing the recall achieved at 90\% precision ($R@P90$). The prediction at $R@P90$ is used as a threshold in the second part of the analysis. We analyse the potential of KBC for missing facts via a manual evaluation.}
  \label{fig:analysis_overview}

\end{figure}

\subsection{Systematic Analysis Procedure}
The goal of our analysis is to realistically assess the abilities of an LM for KBC. Therefore we perform a two-fold analysis by investigating 1) the existing-fact prediction ability of BERT in an automated evaluation and 2) the potential for KBC using LMs by predicting missing facts and evaluating the LM's predictions using human annotations.

The first analysis part includes the automated evaluation of existing facts in \ourdataset{} compared to the LAMA-T-REx benchmark in order to get a realistic estimate of an LM's prediction ability. Based on this evaluation we will extract relation-specific prediction thresholds considering precision and recall trade-offs to enable KBC on high precision and reasonable recall. 

In the second analysis part, we produce high accuracy predictions for missing facts in Wikidata given the previously extracted threshold and evaluate the model's predictions using human annotations. Given the human evaluation, we will provide estimations about the amount of addable new facts to Wikidata. In Figure \ref{fig:analysis_overview} we show an overview of our systematic analysis. 



\section{Datasets}


\subsection{LAMA-T-REx}

The LAMA-T-REx dataset \cite{petroni} is derived from T-REx dataset \cite{trex}, which is a subset of Wikidata and provides alignment between Wikidata (WD) triples and Wikipedia texts. LAMA-T-REx consists of 41 relations with $\sim1000$ facts per relation sampled from T-REx. Included with the dataset are relation-specific, manually defined prompts, which \cite{petroni} also refers to as templates. We use these 41 relations and their corresponding templates throughout our work. The scope of subjects in LAMA-T-REx is restricted to having a Wikipedia page and contains little data as shown in Table \ref{tab:datasets}. This makes it difficult to realistically assess LMs for KBs. 

\subsection{\ourdataset{}}
\label{wd_known}
To realistically assess the usability of LMs for KBC, we created a large-scale dataset with random facts from Wikidata, without subject restrictions to pages such as Wikipedia. One must observe that, while \ourdataset{} is an unbiased sample from Wikidata's subject set per class, it is still biased towards reality, like Wikidata itself \cite{wikidata_bias}.

\textbf{Creation.} We extract facts from Wikidata for the same 41 relations as in LAMA-T-REx. We extract subject-relations pairs per relation and because these pairs may have multiple valid objects, e.g. N:M relations, we extract all associated valid objects along with the pairs. Otherwise, we would risk incomplete ground truth data. This would falsify our evaluation as an LM's prediction can not be recognized as correct when predicting another valid object than the extracted one. We sampled a maximum of $100,000$ subject-relation pairs per relation along with all valid objects. If a relation contains fewer than $100,000$ pairs, we extract all of them. The extracted facts consist of Wikidata-specific entity IDs, which we converted into natural language labels to allow probing an LM for facts. 
In contrast to \cite{petroni}, we do not remove multi-token objects like  ``Ball Aerospace \& Technologies`` because the inability to predict such objects is part of (some) LM's limitations and at the time KBC is performed it is unknown which objects are multi-token objects and which are not as KBC includes predicting missing facts, i.e. facts without ground truth object yet. 
This dataset feature enables the testing of LMs with multi-token prediction capability in comparison to uni-token predictions for KBC. Additionally, a large dataset like \ourdataset{} enables fine-tuning for fact prediction. In Table \ref{tab:datasets} we report some dataset statistics in comparison to LAMA-T-REx showing the larger size of \ourdataset{}. The dataset is available at \url{https://github.com/bveseli/LMsForKBC}.


\vspace{-5mm}

    \begin{table}
        \centering
        \setlength\tabcolsep{5pt}
        
        \scalebox{0.90}{
        \begin{tabular}{l r r r r r r}
        \toprule
        Dataset & &\shortstack{\#unique\\subjects}&\shortstack{\#unique\\objects}&\#triple& \shortstack{\#multi-token\\objects} &\shortstack{object dist. \\ entropy} \\ [1ex]
        \hline
        \multirow{2}{*}{LAMA-T-REx}  & total& 31,479 & 61,85 & 34,039 &0 &-\\
        &  average & 767 & 150 & 830 & 0 & 0.76\\ [1.5ex]
        \multirow{2}{*}{\ourdataset{}} &total&  \textbf{2,956,666}  & \textbf{709,327} & \textbf{3,992,386} & \textbf{1,892,495}&-\\
        & average & \textbf{72,113} & \textbf{17,300} & \textbf{97,375}&  \textbf{46,158}&0.67\\ [1.5ex]
        \hline
        \end{tabular}
        }
        \vspace*{2mm}
        \caption{Our \ourdataset{} dataset in comparison to LAMA-T-REx. We report the total number of distinct objects (\#unique objects), distinct subjects (\#unique subjects), and the number of triples (\#triples) as well as the total number of objects consisting of more than one token (\#multi-token objects), and the average object entropy.
        }
        \label{tab:datasets}
        \vspace{-10mm}
    \end{table}

\section{Potential for Existing-Fact Prediction}
\label{known_facts}
Existing-fact prediction describes the prediction of an object for a subject-relation pair for which the ground truth object already exists in a KB. We will analyse the prediction ability of BERT given existing facts from Wikidata in an automated evaluation, focusing on the recall achieved at 90\% precision ($R@P90$).


\subsection{Metric} We use a rank-based metric and calculate the recall that our method achieves at 90\% precision ($R@P90$).  To compute $R@P90$, we sort the predictions in descending order of their prediction probability, and threshold predictions when average precision drops below 90\%. When determining which prediction is true or false, we have to consider that a subject-relation pair can be associated with multiple valid objects. Therefore, a prediction is true, when it is among the valid objects and false otherwise. 

\subsection{Baselines} \label{baselines} We want to check if BERT's fact prediction ability goes at least beyond just predicting statistically common objects, e.g. English for \texttt{spokenLanguage}. Therefore, we try two distribution-based baselines: random and majority vote. We compute relation-specific object distributions based on a relations ground truth data. In the case of a majority vote, we assign the most probable object to each fact and in the case of random, we assign a random object from the distribution. Additionally, we compare BERT to a relation extraction (RE) baseline and use the \textit{Rosette Text Analytics}\footnote{\url{https://www.rosette.com/capability/relationship-extractor/\#tech-specs}} relation extractor. Given a text snippet, the relation extractor can extract up to 17 relations. For the intersection of these 17 and our 41 relations, we subsampled relation-wise 200 facts from \ourdataset{} and align each fact with text. For text alignments, we consider two different source types: web texts and Wikipedia pages to cover facts with Wikipedia-unknown and Wikipedia-known subjects. Per relation, we align 100 facts with web texts, i.e. top 10 Google search results after googling the subject of a fact and 100 facts with wikipages, i.e. the summary of a subject's Wikipedia page. We get 200 facts with text alignments per relation \texttt{headquarteredIn} and \texttt{citizenOf}, 400 in total.


\subsection{Evaluation and Results} 
\label{eval_results}

    \begin{table}
        \centering
        \setlength\tabcolsep{5pt}
        \scalebox{0.75}{
        \begin{tabular}{ l | l | l || l | l | l }
            \toprule
            & \multicolumn{1}{c}{\ourdataset{}} &  \multicolumn{1}{c}{LAMA-T-REx} & & \multicolumn{1}{c}{\ourdataset{}} &  \multicolumn{1}{c}{LAMA-T-REx}\\
             \cmidrule(lr){2-2} \cmidrule(lr){3-3} \cmidrule(lr){5-5} \cmidrule(lr){6-6}
            
            Relation          & $R@P90$   & $R@P90$       &         Relation           & $R@P90$   & $R@P90$     \\
            \midrule
            nativeLanguage             &  \textbf{0.61}     &  0.68   &        foundedIn              &  0.00009  & 0.001   \\
            spokenLanguage                &  \textbf{0.26}     & 0.33   &        deathPlace              &  0.00009  & 0  \\
            officialLanguage            &  \textbf{0.25}     & 0.37   &        namedAfter             &  0.00008  & 0      \\
            headquarteredIn             &  \textbf{0.04}     & 0.52    &        partOf                 &  0.00006  & 0   \\
            developedBy                &  \textbf{0.04}     & 0.64    &        twinTown                &  0.00003  & 0.001   \\
            producedBy                  &  \textbf{0.03}     & 0.86    &        sharesBorders         &  0.00001  & 0.001   \\
            countryOfJurisdiction     &  \textbf{0.02}     & 0.66    &        fieldOfWork            &  0        & 0     \\
            hasCapital                 &  \textbf{0.01}     & 0.60    &        employedBy              &  0        & 0.002 \\  
            locatedIn                   &  \textbf{0.008}    & 0.20    &        hasReligion   &  0        & 0         \\
            bornIn                      &  0.006    & 0.009    &        playerPosition         &  0        & 0         \\
            isCapital                  &  \textbf{0.006}    & 0.81   &        subclassOf           &  0        & 0.01         \\
            CountryOfOrigin             &  0.005    & 0.08    &        holdsPosition        &  0        & 0         \\
            isA                         &  0.004    & 0.06   &        diplomaticRelation    &  0        & 0         \\
            LanguageOfFilm              &  0.003    & 0  &        citizenOf             &  0        & 0         \\
            ownedBy                 &   \textbf{0.0008}  &    0.16 &        consistsOf            &  0        & 0         \\
            hostCountry                  &  0.0006   & 0.002 & musicGenre             &  0        & 0         \\
            originalBroadcoaster      &  0.0004   & 0.02  &        musicLabel             &  0        & 0.02         \\
            inTerritoryOf            &  0.0002   & 0.01  &        playsInstrument     &  0        & 0.003         \\
            writtenIn                &  \textbf{0.0001}   & 0.15  &        hasProfession         &  0        & 0         \\ 
            locationOfWork          &  0.0001   & 0  &       inContinent           &  0        & 0.004         \\
            memberOf                &  \textbf{0.0001}   & 0.52 &      & & \\  
            \bottomrule
        \end{tabular} 
        }
        \vspace*{2mm}
        \caption{BERT's performance on data sets \ourdataset{} a LAMA-T-REx for the same 41 relations. Boldface marks significantly lower values, indicating an overestimation of BERT's fact predicting ability on LAMA.}
        \label{tab:r_at_pk_table}
        
    \end{table}

\noindent
\subsubsection{Quantitative Analysis.} In Table \ref{tab:r_at_pk_table} we report $R@P90$ values achieved by BERT's predictions on our dataset \ourdataset{} in comparison to \textit{LAMA-T-REx}. On \ourdataset{} the LM achieves significantly lower values (marked in bold) suggesting a more realistic assessment of BERT's fact prediction ability by \ourdataset{}. Only the relations  \texttt{nativeLanguage, spokenLanguage,off\-icialLanguage} show a smaller decrease and therefore stable results. 

    \vspace{-6mm}
    \begin{figure}
      \centering
      \includegraphics[width=85mm]{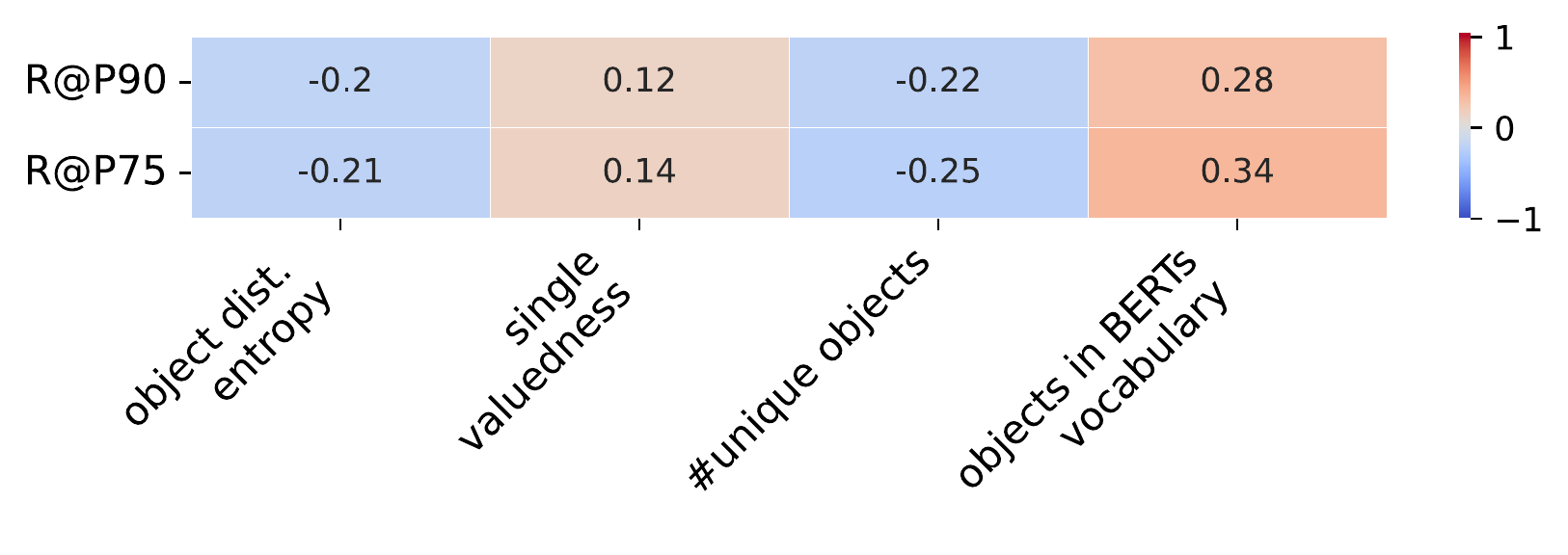}
      \vspace{-5mm}
      \caption{Pearson Correlation Analysis}
      \label{fig:corr_anal}
    \end{figure}
    \vspace{-6mm}

\noindent
To investigate why BERT achieves low results, we perform a correlation analysis computing the Pearson correlation coefficient over $R@P90$ in Figure \ref{fig:corr_anal}. We notice a negative correlation between $R@P90$ and the object distribution entropy, meaning that a more uniform distribution makes the predictions harder. Furthermore, the number of unique objects is also negatively correlated with BERT's performance, i.e. fewer possible objects benefit the performance. The single valuedness, i.e. the relation-wise proportion of subject-relation pairs with only one object, shows a low but positive correlation. This indicates the performance is better on N:1 or 1:1 relations confirming the observation in \cite{petroni}. The performance is also positively correlated with the number of objects in BERT's vocabulary, i.e. the more objects of a relation are covered by the LM's vocabulary, the better the performance. The vocabulary acts as a bottleneck, preventing the model from predicting facts.

Looking at the baselines in Table \ref{tab:rosette_baseline}, we see that the majority baseline is quite solid with an average precision of 0.18. Access to the underlying distribution of relation-specific ground truth data has a noticeable impact on assigning objects to a subject-relation pair. Still, BERT achieving an average precision in a higher range ($>75\%$) shows that the prediction ability is based on more than predicting statistically common objects. The RE baseline is outperformed by BERT for two tested relations, while BERT and RE show lower results on facts aligned with webtexts than on facts aligned with Wikipedia pages.

    \begin{table}
        \centering
        \setlength\tabcolsep{5pt}
        
        \scalebox{0.85}{
        \begin{tabular}{l ccc || l ccc ccc}
            \toprule
            \multicolumn{4}{c}{\shortstack{Distribution vs. BERT}} &  \multicolumn{7}{c}{RE vs. BERT}\\
            \cmidrule(lr){1-4} \cmidrule(lr){5-11}
            &&&&& \multicolumn{3}{c}{\shortstack{Wikipedia}} &  \multicolumn{3}{c}{Google search}\\
             \cmidrule(lr){6-8} \cmidrule(lr){9-11} 
            &$\overline{P}$ & $\overline{R}$ &$F1$& & $\overline{P}$   & $\overline{R}$  & $F1$  & $\overline{P}$  & $\overline{R}$    & $F1$ \\
            \midrule
            Random & 0.09 & 0.05 & 0.06& Rosette & 0.32 & 0.05 & 0.08 & 0.21 &0.01 & 0.02 \\
            Majority & 0.18 & 0.09  & 0.03& BERT @P75 & 0.87 & 0.31 & 0.45 & 0.37 & 0.28 & 0.31 \\
            BERT @P75 & 0.75 & 0.43 & 0.54&  BERT @P90 & 0.95 & 0.22 & 0.17 & 0.45 & 0.23 & 0.30 \\
            BERT @P90 & 0.90 & 0.35 & 0.48&  & & & & &\\
            \bottomrule
            
        \end{tabular}  
        }
        \vspace*{2mm}
        \caption{Random and Majority Baselines (left) and Relation Extraction (RE) Baseline for \texttt{citizenOf} and \texttt{headquarteredIn} (right). The RE Baseline is done on two datasets: 1) a (Wikipedia) dataset, where the triples are aligned with the Wikipedia summaries extracted from the subject's Wikipedia pages, and 2) a (Google search) dataset, where the triples are aligned with texts from the top 10 Google search results after searching for the subject of the respective triple. Scores were computed on a subset of \ourdataset{} with text alignments as described in \ref{baselines}.}
        \label{tab:rosette_baseline}
        \vspace{-7mm}
    \end{table}

\subsubsection{Qualitative Analysis.} 

Since we are interested in high precision for KB completion and quantitative analysis showed low R@P90 values for most of the relations, we need to increase BERT's performance. We first perform a qualitative analysis of issues. Since analysing all 41 relations qualitatively is not feasible, we select a representative subset of relations. The subset is diverse regarding, e.g. semantics (e.g. language, demographic, goods), entity type (human, company, places), performance (lower vs. higher scores), and possible objects (all languages vs. cities worldwide). The chosen relations are shown in Table \ref{tab:main_results}. We aim to identify and eliminate or at least reduce systematic prediction errors for those relations. Common error categories are: 1) predicting the wrong hierarchy, i.e. country instead of a city; 2) predicting the wrong category, i.e. country instead of nationality and 3) ambiguous prompts, i.e. predicting ``local`` for ``Albert Einstein is a [MASK] citizen``. Such cases falsify the evaluation, since a distinction between actually true predictions (``German`` or ``Germany``) and actually false predictions (``local`` or ``German``) is not made. These errors are rooted in the conceptual discrepancy between LMs and KBs. While querying a KB is clearly defined and leads to unambiguous answers, an LM is queried by natural language prompts, which are as ambiguous as natural language itself. To use LMs for KBs we need to translate between them. Therefore we focus on three main model parts: input, model, and output: 1) input optimization by adjusting prompts; 2) model optimization by fine-tuning, and 3) output adjustment by converting ground truth and prediction into the same category. Input and model optimizations are done on relation-wise data splits for training and test $(80-20)$, where we split based on subject-relation pairs avoiding having the same subject in training and test.

\textit{Input Optimization.} \textit{AutoPrompt} \cite{autoprompt} introduced an automatic way of prompt generation given a specific task, such as fact retrieval. We generate our prompts as suggested by \cite{autoprompt} on relation-specific training splits of \ourdataset{}. In Table \ref{tab:main_results} the input optimization with AutoPrompts achieves notable improvements for $R@P90$, e.g. \texttt{hasReligion} increased from $0$ to $0.21$, \texttt{citizenOf} from $0$ to $0.15$. There is no deterioration in any of the relations. 

\textit{Model Optimization.} Fine-tuning not only allows us to optimize an LM on the searched output but also enables adding new words to the model's vocabulary as this has been shown to be a bottleneck for fact prediction. We fine-tune relation-specific LMs and for vocabulary extension, we add a $max.$ of $2000$ objects per relation. For fine-tuning the masked language model objective \cite{bert} is used as well as our training splits. We show results on two setups: 1) fine-tuning $BERT_{large}$, denoted  $FT$, and 2) fine-tuning $BERT_{large}$ and with extended vocabulary, denoted $FT_{vocab}$. In Table \ref{tab:main_results} the model optimization shows the biggest improvements. Only \texttt{developedBy, producedBy, headquarteredIn} deteriorate at first in the $FT$ setup, but \texttt{producedBy} and \texttt{developedBy} improve significantly in the $FT_{vocab}$ setup. We found that after fine-tuning the model predicts substrings that match the ground truth object, e.g. in relation \texttt{producedBy} ``Len`` is predicted if the ground truth object is ``Lenovo`` or ``Xiao`` for ``Xiaomi``. The same happens in relation \texttt{developedBy}, e.g. ``Core`` for ``Corel``, and in relation \texttt{headquarteredIn} e.g. ``Wind`` for the city ``Windhoek``. After vocabulary expansion previously missing tokens like ``Xiaomi`` can now be fully predicted, so that $R@P90$ can increase in the $FT_{vocab}$ setup. It is worth noting that  \texttt{producedBy} and \texttt{developedBy} could only be improved significantly by expanding the vocabulary during fine-tuning. Only for \texttt{headquarteredIn} the precision does not improve. While \texttt{headquarteredIn} does degrade in precision, fine-tuning with an extended vocabulary increases the overall quality significantly (see Table \ref{tab:main_results}, AVG $R@P90$).

\textit{Output Adjustment.} To map prediction and ground truth, so a prediction ``French`` is correctly recognized for object ``France``, we use manually crafted dictionaries. Methods like string matching can always lead to incorrect mappings and sometimes even do not work for examples like the one shown. Therefore, we used two dictionaries mapping nationalities and countries on the one hand and religions and religious affiliations on the other hand. A prediction will be true if it belongs to the same entry in a relation-specific dictionary as the ground truth object and false otherwise. 
The creation of such mapping dictionaries involves tremendous manual labor, contradicting our search for automated KBC.
Therefore, we evaluate on only two \texttt{hasReligion} and \texttt{citizenOf} as these relations were also most affected by the second error category mentioned above so that the output adjustment here might have the greatest effect. We find, that automated fine-tuning significantly outperforms this approach.

\begin{table}
    \centering
    \setlength\tabcolsep{5pt}
    \scalebox{0.75}{
    \begin{tabular}{ll cc cc}
        \toprule
        & \multicolumn{1}{c}{Base} &  \multicolumn{1}{c}{ \shortstack{Input\\ Opt.}} & \multicolumn{2}{c}{Model Opt.} &  \multicolumn{1}{c}{\shortstack{Output \\ Adjustment}} \\
         \cmidrule(lr){2-2} \cmidrule(lr){3-3} \cmidrule(lr){4-5} \cmidrule(lr){6-6}
        Relation        & Pre-Trained   & AutoPrompt   & $FT$   & $FT_{vocab}$   & \shortstack{Manual \\Mapping} \\
        \midrule
        nativeLanguage  &   0.62        & 0.66  & 0.79  & \textbf{0.79}          & -        \\
        hasReligion     &   0           & 0.21  & 0.13  & \textbf{0.27}          & 0.02        \\
        citizenOf       &   0           & 0.15  & 0.23  & \textbf{0.23}         & 0.01        \\
        producedBy      &   0.03        & 0.03  & 0     & \textbf{0.15}          & -         \\
        developedBy     &   0.04        & 0.04  & 0.0004& \textbf{0.11}           & -         \\
        headquarteredIn &   \textbf{0.04}        & 0.04  & 0     & 0               & -         \\
        spokenLanguage    &   0.26        &0.42   & \textbf{0.51}  & 0.5            & -         \\
        LanguageOfFilm  &   0.003     & 0.04  & \textbf{0.29}  & 0.27         & -         \\
        \midrule[\heavyrulewidth]
        AVG $R@P90$       &0.12           &0.20   &0.24   &\textbf{0.29}              & 0.015\\
        \bottomrule
        
    \end{tabular}  
    }
    \vspace*{2mm}
    \caption{Table shows $R@P90$ values. Improvement approaches to maximize BERT's performance on specific relations in comparison to pre-trained BERT-large (case-sensitive). Improvements were tested at three key points in the LM: \textit{input, model, output}. Scores were computed on the test split of \ourdataset{}.}
    \label{tab:main_results}
\end{table}

\subsubsection{Summary.} We found that using biased datasets lead to an overestimation of an LM's performance for fact prediction. To neither over- nor underestimate BERT's performance, we tested it on the large dataset \ourdataset{} and implemented improvements to increase BERT's performance and perform KBC with high precision later on. We have seen the model's vocabulary to be a bottleneck for its performance. When fine-tuning with vocabulary extension, the model's performance can be significantly improved for fact prediction.



\section{Potential for Knowledge Base Completion}
\label{unknown_facts}

In the following, we will obtain plausible missing facts from Wikidata, produce high accuracy predictions respective to $R@P90$, and let annotators from Amazon Mechanical Turk (mturk) manually evaluate the predictions given a five-value scale \textit{true, plausible, unknown, implausible, false}. The relations in focus are the same as in Table \ref{tab:main_results}, except for \texttt{hasReligion}, which appeared with too sparse information on the web.

\subsection{Human Evaluation} 
\label{data} 

\subsubsection{Obtaining missing facts.}
\label{wd_unknown}

The key to KBC with LMs is the ability to predict \textit{missing facts}. 
Directly extracting empty subject-relation pairs from Wikidata is not possible, since Wikidata consists of existing facts $(s,r,o)$. Also, randomly sampling an arbitrary subject to combine it with any relation would run the risk of violating our condition of plausible missing facts, where an object for a subject-relation pair exists (e.g. an implausible pair like (Albert Einstein, hasCapital,$\cdot$) has no object). Therefore, we will only sample subject-relation pairs, where the subject has a relation-compatible entity type. Thus, we compute relation-wise the most frequent subject entity type within a relation. When sampling subject-relation pairs with missing objects, the subject is conditioned on having the most frequent entity type. This ensures extracting plausible missing facts.  We randomly sample $10,000$ missing facts per relation, $70,000$ missing facts in total.


\subsubsection{High Accuracy Predictions.} 

    To provide human annotators with reasonable predictions, we set a relation-specific prediction threshold to ensure the prediction quality and use the best possible model for predictions based on results in Table \ref{tab:main_results}. Given these best models, the threshold is the prediction probability at $R@P90$ of each relation to respect the relation-specific precision and recall trade-offs. We keep only those missing facts, i.e. subject-relation pairs, whose predictions have a probability over the threshold. These are our \textit{high accuracy predictions}.

\subsubsection{Annotations.}
\label{eval_unknown}
    We filter the $70,000$ missing facts relation-wise by keeping only the facts with high accuracy predictions and then sample $50$ missing facts with high accuracy predictions. This results in one prediction per subject-relation pair, i.e. 350 predicted missing facts in total.
    For these 350 missing facts, we use Amazon Mechanical Turk (mturk) for human annotations.  An annotator is asked to evaluate a predicted missing fact. For readability reasons, each fact is formulated into relation-specific statements such as ``The native language of Marcus Adams is English``, where ``English`` is the prediction and \texttt{nativeLanguage} the relation. The annotators are given a five-value scale: true, plausible, unknown, implausible, and false. Before voting, we ask them to look up the fact on the web. They are required to give an evidence link and copy a text snippet supporting their voting. In case they vote unknown, they have to explain their voting. This way we ensure that annotators leave reasonable votes. We can say that all annotators left understandable insights regarding their voting.
    
    We also self-annotated the 350 missing facts with ground truth and evidence links. Along with ground truth annotations, we annotated if the subject was known to English (en) Wikipedia; if the ground truth consists of more than one word; if the ground truth is in BERT's vocabulary, the position of the found evidence in Google search results and the search link used to find the ground truth. Furthermore, we rated each prediction using the same five-value scale as mturk annotators. We and mturk annotators reach an agreement of $69\%$ given the five-value scale and an agreement of $94\%$ given the upper categories true (true, plausible) and false (unknown, implausible, false). 


    \begin{table}
        \centering
        \setlength\tabcolsep{5pt}
        \scalebox{0.75}{
        \begin{tabular}{l | c | c @{\hskip 0.2in}||@{\hskip 0.2in} c | c | c | c | r |c }
            \midrule
            Relation            & true & false  &  \shortstack{true with\\ evidence} &  plausible  & unknown & implausible  & \shortstack{false with\\evidence} & \shortstack{in Wikipedia\\ (en)}  \\
            \midrule
            nativeLanguage      & $82\%$ & $18\%$ &  $48\%$ & $34\%$ & $6\%$ & $8\%$ & $4\%$ & 16\% \\
            spokenLanguage         & $82\%$ & $18\%$ & $46\%$ & $36\% $&$6\%$ & $6\%$& $6\%$ & 28\%\\
            headquarteredIn      & $82\%$ & $18\%$ &  $34\%$ & $48\%$ & $6\%$ & $8\%$ & $4\%$ & 26\%\\
            developedBy           & $62\%$ & $38\%$ & $50\%$ & $12\% $& $0\%$ & $0\%$ & $38\%$ & 74\% \\
            producedBy           & $22\%$ & $78\%$ & $20\%$ & $2\%$ & $6\%$ & $0\%$ & $72\%$ & 10\%\\
            LanguageOfFilm       & $76\%$ & $24\%$ & $56\%$ & $20\%$ &$0\%$ & $2\%$ & $22\%$ & 32\%\\  
            citizenOf           &  $90\%$ &  $10\%$ & $52\%$ & $38\%$ & $4\%$ & $0\%$ & $6\%$  & 8\%\\
            \bottomrule
        \end{tabular}  
        }
        \vspace*{2mm}
        \caption{Overview of the results from the human evaluation. ``True`` denotes the summed up human ratings for ``true with evidence`` and ``plausible`` per relation. Similarly, ``false`` denotes the combined human ratings for ``unknown``, ``implausible``, and ``false with evidence`` per relation. The column ``in Wikipedia (en)`` describes the ratio of subjects with English Wikipedia articles to subjects without English Wikipedia articles per relation. }
        \label{tab:human_annot}
    \end{table}

\vspace{-12mm}
\subsection{Results} 
\label{kbc_results}

\textbf{Quantitative results.} In Table \ref{tab:human_annot}, we see that the human annotators rate the model's predictions as highly correct. Based on these values we have determined the potential for KBC in Table \ref{tab:kbc}. Given the number of missing facts and the proportion of high accuracy predictions, we can estimate the amount of addable facts at a relations-specific accuracy. This accuracy was achieved through human evaluation as shown in Table \ref{tab:human_annot}. Given the relation \texttt{nativeLanguage} we could add $5,550,689$ new facts in a human-in-a-loop procedure or $7,871,085 \cdot 0.86=6,769,133$ with an accuracy of $82\%$ automatically. In a human-in-a-loop procedure, the estimated costs of 2 USD (section \ref{intro}) for manually curated facts could drop to 40 Cents with approximately 2 minutes per annotation as we experienced with our mturk evaluation. Given our results in Table \ref{tab:kbc} we can perform significant KBC for relations \texttt{nativeLanguage} and \texttt{spokenLanguage} at precision 82\% and \texttt{citizenOf} at precision 90\%. 


\textbf{Qualitative results.} 
Looking into the annotations and predictions, we see that statements such as ``Final Fantasy VII is developed by Square`` are almost literally included in corresponding evidence links such as  Wikipedia pages\footnote{\url{https://en.Wikipedia.org/wiki/Final\_Fantasy\_VII}}. In contrast, relations like \texttt{nativeLanguage} include statements only implicitly, e.g. the native language of ``Marcus Adams`` is never mentioned explicitly in the corresponding Wikipedia page\footnote{\url{https://en.Wikipedia.org/wiki/Marcus\_Adams\_(Canadian\_football)}}.
Yet, the LM achieves comparable results on most relations despite their more implicit or explicit factual basis. 

\textbf{Generalization or Retrieval?} To investigate this further, we computed the proportion of subjects with English Wikipedia articles per relation. This enables us to estimate whether facts were mentioned in corresponding Wikipedia articles and thus, were present in BERT's training set. It can be shown that the language-related and socio-demographic relations achieve high results despite their lower occurrence in Wikipedia. This means that BERT predicts never seen facts with a high accuracy for these relations, showing a high generalization capability. When given facts such as the headquarters of  ``Universal de Televisión Peru``, the model correctly predicts ``Lima``.
Or given a fact such as the original language of the movie ``Il mio paese``, the model predicts ``Italian`` correctly.
Socio-demographic relations such as \texttt{citizenOf}, \texttt{headquarteredIn} or language-related relations like \texttt{nativeLanguage}, \texttt{LanguageOfFilm} exhibit stronger correlations (e.g. person name and spoken language/origin country) than other relations (e.g. video game and developer, goods/products and manufacturer). These correlations are learned by the LM from the vast amounts of training data and used for the prediction. We recognize here that such learned correlations can be quite useful for fact prediction.
Regarding the non-language or non-socio-demographic relations, we see that \texttt{producedBy} has the least Wikipedia-known subjects of 10\% and shows also the lowest accuracy of 22\%. In contrast, \texttt{developedBy} has the most Wikipedia-known subjects and still a high accuracy of 62\% despite being a non-language or non-socio-demographic relation. In these relations, the model shows less generalization capability and is more in need of actual retrieval. As an example: the developer of ``Pro Cycling Manager 2015`` must be explicitly mentioned during training to know it, yet the model correctly predicts ``Cyanide``. 

\textbf{Conclusion.} Given these qualitative examples and quantified numbers the model is capable of generalization as well as retrieval. But it is still unclear in what mixture and to what extent for example fact retrieval is possible. Regarding KBC both are beneficial. In case of precise retrieval, facts are addable automatically to an existing KB. In the case of generalization, which still achieves a high accuracy, human-in-a-loop procedures allow adding manually curated facts in a faster and cheaper way.

    \vspace{-3mm}
    \begin{table}
        \centering
        \setlength\tabcolsep{5pt}
        \scalebox{0.75}{
        \begin{tabular}{l | r | r | r | c | r | r  }
            \midrule
            Relation  & $cardinality^{WD}$ & \shortstack{\#missing \\facts} & \shortstack{high accuracy\\ predictions$(\%)$} & accuracy &\shortstack{\#addable \\facts} & \shortstack{growth\\factor} \\ 
            \midrule
            nativeLanguage  &  264,778 &  7,871,085 & 86\% & 82\% & 5,550,689 & 20.96 \\
            spokenLanguage    &   2,148,775  & 7,090,119 & 77\% & 82\% & 4,476,701 & 2.08 \\
            headquarteredIn      & 409,309 & 55,186 & 8\% & 82\% & 3,443 & 0.008 \\
            developedBy       & 42,379 & 29,349 & 2\% & 62\% & 363 & 0.01 \\
            producedBy          & 123,036& 31,239 & 0.8\% & 22\% & 55 & 0.0004 \\
            LanguageOfFilm      &   337,682  & 70,669 & 37\% & 76\% & 19,872 & 0.06 \\
            citizenOf       &     4,206,684 & 4,616,601 & 28\% & 90\% & 1,163,383 & 0.27 \\
            \bottomrule
        \end{tabular}  
        }
        \vspace*{2mm}
        \caption{The amount of missing facts and the percentage of high accuracy predictions denotes the number of new facts we could add at a relation-specific precision. The amount of addable facts indicates the number of potential new facts that could be added without error, e.g. in a human-in-a-loop procedure. The growth factor describes the potential growth of Wikidata given the current $cardinality^{WD}$ in Wikidata and the amount of addable facts.}
        \label{tab:kbc}
        \vspace{-10mm}
    \end{table}

\section{Discussion and Conclusion}

In this paper, we investigated the potential of automated KB completion using LMs. We introduced a challenging benchmark dataset, \ourdataset{}, an unbiased random sample of Wikidata containing 3.9M existing facts. Using this dataset enabled a more realistic assessment of KB completion using LMs. This revealed that previous benchmarks lead to an overestimate of LM-based KB completion performance. 

Our analysis showed that LMs are not able to obtain results at a high precision ($\sim90\%$) for all relations equally, but LM-based knowledge covers language-related and socio-demographic relations particularly well. Furthermore, we discovered that an LM's vocabulary can limit the capability of fact prediction and we achieved significant improvements with fine-tuning and vocabulary expansion.

Since the prediction of facts non-existent to the KB is crucial for KB completion, we extracted plausible subject-relation pairs with non-existent objects in the KB. By probing the LM for these facts, we received actual novel facts previously unknown to the KB. Since the ground truth for these facts is missing, we performed a human evaluation. Annotators rated the LM's suggestions as highly correct. That showed a high potential for KB completion, either completely automated at a precision of up to $90\%$ or as a strong recommender system for human-in-a-loop procedures. We demonstrated that in a human-in-a-loop procedure, LMs might reduce the costs for manually curated facts significantly, from approximately \$2 to \$0.4 per fact.

Moreover, we showed that LMs build surprisingly strong generalization capabilities for specific socio-demographic relations.

A promising direction for future work could be the construction of LMs specifically for KBs, which goes beyond fine-tuning. This could include defining specific vocabularies that are optimized for fact prediction.


\bibliographystyle{splncs04}
\bibliography{references}

\end{document}